\documentclass[journal,twoside,web]{ieeecolor}
\usepackage{tmi}
\usepackage{cite}
\usepackage{amsmath,amssymb,amsfonts}
\usepackage{algorithmic}
\usepackage{graphicx}
\usepackage{textcomp}

\usepackage{makecell}
\usepackage{multirow}
\usepackage{booktabs}
\usepackage{tabularx}
\usepackage{ragged2e} 
\usepackage{caption}
\usepackage{pifont}
\usepackage{marvosym}
\usepackage{ifsym}
\usepackage[table]{xcolor}

\definecolor{iccvblue}{rgb}{0.21,0.49,0.74}
\definecolor{myblue}{RGB}{0, 138, 218}
\definecolor{myorange}{RGB}{218,138,0}
\usepackage[colorlinks=true, linkcolor=blue, citecolor=blue, urlcolor=iccvblue]{hyperref}
\usepackage{CJKutf8}

\def\BibTeX{{\rm B\kern-.05em{\sc i\kern-.025em b}\kern-.08em
    T\kern-.1667em\lower.7ex\hbox{E}\kern-.125emX}}
\begin{document}
\title{PathMR: Multimodal Visual Reasoning for Interpretable Pathology Diagnosis}
\author{
Ye Zhang, Yu Zhou, Jingwen Qi, Yongbing Zhang,~\IEEEmembership{Senior Member,~IEEE}, 
Simon P\"uttmann, Finn Wichmann, Larissa Pereira Ferreira, Lara Sichward, Julius Keyl, Sylvia Hartmann, Shuo Zhao, Hongxiao Wang, Xiaowei Xu, Jianxu Chen
\thanks{
Ye Zhang, Yu Zhou and Jingwen Qi contributed equally to this work.  
This work was supported in part by the National Natural Science Foundation of China (62031023, 62331011), 
the Federal Ministry of Education and Research in Germany (Funding Reference 161L0272), 
and the Ministry of Culture and Science of the State of North Rhine-Westphalia. 
\textrm{\Letter}~Corresponding authors: Yongbing Zhang (ybzhang08@hit.edu.cn) and Jianxu Chen (jianxu.chen@isas.de).
}
\thanks{
Ye Zhang and Yongbing Zhang are with the School of Computer Science and Technology, Harbin Institute of Technology, Shenzhen 518055, China (zhangye94@stu.hit.edu.cn).
Yu Zhou, Simon P\"uttmann, Shuo Zhao, Xiaowei Xu, and Jianxu Chen are with the Leibniz-Institut für Analytische Wissenschaften – ISAS – e.V., Dortmund 44139, Germany (yu.zhou@isas.de, simon.puettmann@isas.de, shuo.zhao@isas.de, xiao.wei.xu@foxmail.com).
Jingwen Qi is with the Department of Pathology, The Sixth Affiliated Hospital, Sun Yat-sen University, Guangzhou 510655, China (qijw6@mail.sysu.edu.cn).
Finn Wichmann, Larissa Pereira Ferreira, Lara Sichward, Julius Keyl, and Sylvia Hartmann are with the Institute of Pathology, University Hospital Essen, Essen 45147, Germany (finnalexander.wichmann@uk-essen.de, larissa.pereiraferreira@uk-essen.de, larahelena.sichward@uk-essen. de, julius.keyl@uk-essen.de, sylvia.hartmann@uk-essen.de).
Julius Keyl is also with the Institute for Artificial Intelligence in Medicine, University Hospital Essen, 45131, Germany.
Hongxiao Wang is with the Academy for Multidisciplinary Studies, Capital Normal University, Beijing 100048, China (hongxiao.wang@isas.de).
Xiaowei Xu is also with the Department of Cardiovascular Surgery, Guangdong Provincial People’s Hospital, Guangdong Academy of Medical Sciences, Southern Medical University, Guangzhou 510080, China.
}
}

\maketitle

\begin{abstract}

Deep learning based automated pathological diagnosis has markedly improved diagnostic efficiency and reduced variability between observers, yet its clinical adoption remains limited by opaque model decisions and a lack of traceable rationale. To address this, recent multimodal visual reasoning architectures provide a unified framework that generates segmentation masks at the pixel level alongside semantically aligned textual explanations. By localizing lesion regions and producing expert style diagnostic narratives, these models deliver the transparent and interpretable insights necessary for dependable AI assisted pathology. Building on these advancements, we propose PathMR, a cell-level \textbf{M}ultimodal visual \textbf{R}easoning framework for \textbf{Path}ological image analysis. Given a pathological image and a textual query, PathMR generates expert-level diagnostic explanations while simultaneously predicting cell distribution patterns. To benchmark its performance, we evaluated our approach on the publicly available PathGen dataset as well as on our newly developed GADVR dataset.
Extensive experiments on these two datasets demonstrate that PathMR consistently outperforms state-of-the-art visual reasoning methods in text generation quality, segmentation accuracy, and cross-modal alignment. These results highlight the potential of PathMR for improving interpretability in AI-driven pathological diagnosis. The code will be publicly available in \href{https://github.com/zhangye-zoe/PathMR}{https://github.com/zhangye-zoe/PathMR}.
\end{abstract}


\IEEEPARstart{A}dvancements in tissue sectioning and microscopic imaging have enabled the digitization of whole slide images (WSIs), laying the foundation for computational pathology. Currently, H\&E-stained pathological image analysis is widely regarded as the gold standard for cancer diagnosis \cite{claudio2024mapping}, and prognostic assessment \cite{fan2022cancer}. Meanwhile, deep learning-based cancer diagnosis models have improved pathological evaluations' efficiency, consistency, and scalability, offering significant advantages over traditional cancer diagnosis approaches.

\begin{figure}[t!]
    \centering
    \includegraphics[width=3.5in]{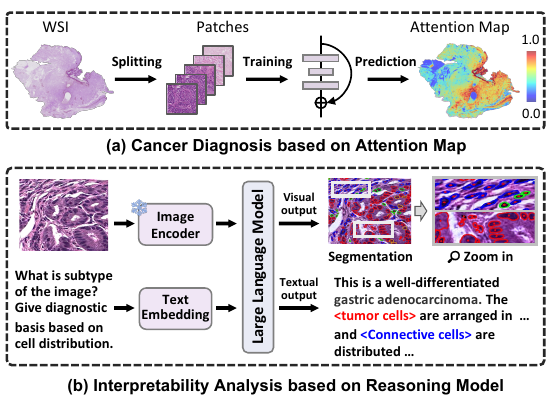}
    \caption{\textbf{Comparison of interpretability analysis methods in pathological diagnosis.} (a) is classification task based on multiple instance learning, which employs an attention mechanism to compute patch importance scores. (b) represents multi-modal visual reasoning model that simultaneously support visual (segmentation) and textual (diagnosis report) outputs.}
    \vspace{-0.4cm}
    \label{fig1}
\end{figure}

Despite these advancements, most automated diagnostic methods have primarily focused on improving accuracy while overlooking the diagnostic interpretability for clinical decisions. Although some multiple instance learning (MIL) approaches introduce an attention mechanism to highlight diagnostically relevant regions \cite{shao2021transmil, zhang2022dtfd} by computing a global attention map as shown in Fig. \ref{fig1} (a). However, these methods fail to capture tumor immune characteristics due to patch-level granularity. \textbf{First}, each patch typically contains hundreds of cells with diverse morphologies and spatial arrangements, making it challenging to capture fine-grained details. \textbf{Second}, attention scores reflect relative importance but do not provide explicit textual explanations, limiting their utilization in clinical decision-making.
In contrast, pathologists usually assess tumor progression and differentiation by identifying tumor regions and microscopic features, such as cell morphology and their spatial distribution. These cellular-level characteristics provide essential insights into the tumor immune microenvironment, which is critical for effective diagnosis and treatment planning. Therefore, developing a fine-grained interpretable model that integrates visual and textual information is crucial for pathology diagnosis.

Recently, in natural scene understanding, visual reasoning models such as LISA and Groma have integrated large language models with specialized visual perception modules to produce segmentation masks at the pixel level and semantically aligned text in one unified process. Each $<$seg$>$ token is explicitly tied to a corresponding image region, ensuring that generated descriptions map precisely to visual features. This synchronous generation framework extends seamlessly to medical diagnosis. As shown in Fig. \ref{fig1} (b), the visual branch accurately delineates lesion areas while the language branch generates expert level diagnostic reports. Through careful modular design, masks and narratives remain tightly coupled both spatially and semantically, providing decision support for AI assisted pathology that is transparent and traceable.

Despite these advances, current visual reasoning frameworks often produce fragmented segmentation outputs, evident as shape distortions and boundary noise. These artifacts stem from misalignment between visual and textual features, which allows spurious predictions to persist. To overcome this, we introduce a novel constraint mechanism in the visual perception module that enforces morphological consistency and suppresses isolated noisy predictions. By applying this mechanism, our model yields cleaner, more coherent masks that align more closely with generated text, enhancing interpretability and reliability in pathological image analysis.

This paper presents PathMR, a cell level multimodal visual reasoning model resilient to fragmentation, designed for interpretable pathology diagnosis. Given a histopathology image and an associated text query, PathMR produces expert level narrative responses alongside cell segmentation and classification masks, ensuring precise semantic alignment between visual and textual outputs. To enable standardized evaluation and foster further research, we introduce GADVR, the first pixel level visual reasoning dataset in pathology, which comprises approximately 190,000 image patches annotated by gastric adenocarcinoma subtype and over 550,000 paired image and text entries. Finally, to address fragmentary segmentation in pixel level reasoning tasks, we incorporate class supervision and consistency constraints that reduce uncertainty in cell category predictions and preserve the integrity of cell morphology.

In general, our contributions are summarized as follows:  

\begin{itemize}
    \item [(1)] We present a novel pixel-level visual reasoning model for pathological image analysis, offering fine-grained cell-level interpretability to improve clinical diagnosis.
    \item [(2)] To facilitate evaluation, we develop a comprehensive pixel-level multimodal visual reasoning dataset, which includes pixel-wise nuclei segmentation and classification labels alongside cell-level question-answer pairs.
    \item[(3)] We propose a dual constraint mechanism that combines classification supervision and a morphological consistency constraint to stabilize cell class predictions and reduce boundary noise, improving the accuracy of pixel level segmentation in visual reasoning models.
    \item [(4)] Extensive experiments across various visual reasoning models demonstrate that our approach consistently outperforms existing state-of-the-art methods in segmentation accuracy and text generation quality.
\end{itemize}

\section{Related Work}  

\subsection{Interpretable Pathological Diagnosis Models}

Deep learning has substantially advanced automated pathological image analysis \cite{swanson2023patterns, wang2024pathology, deng2024cross}. However, its clinical integration remains limited, largely due to the lack of interpretability and transparency in model decision-making. To address this, attention-based methods have been developed to provide patch-level interpretability by assigning importance scores to image regions. In particular, multiple instance learning (MIL) frameworks \cite{li2021dual, zhang2022dtfd, shao2021transmil} utilize attention mechanisms to highlight diagnostically relevant lesion areas. While effective in localizing salient regions, these approaches often produce coarse attention maps that lack clinical validation and do not reflect the reasoning process of pathologists.
To move beyond attention, multimodal strategies have been proposed to enhance interpretability by aligning image features with textual information. For instance, Zhang et al. \cite{zhang2019pathologist} linked histopathology patches to textual descriptions, enabling a more human-understandable model output. Others have focused on quantifying cellular-level feature distributions across tumor stages \cite{jaume2021quantifying, zamanitajeddin2024social}, aiming to provide statistical insights into disease progression. Despite these advances, most existing methods still lack fine-grained interpretability, highlighting the importance of developing interpretability frameworks that operate at the microscopic level to reflect clinical reasoning.

\subsection{Multimodal Pathology Models} 
Early vision language efforts in computational pathology demonstrated that gigapixel scale WSIs can be used for downstream question answering and report generation. Chen et al. introduced WSI-VQA, the first framework to perform slide level visual question answering \cite{chen2024wsi}, and WSICaption, which generates diagnostic captions directly from WSIs\cite{chen2024wsicaption}, proving that transformer based models can attend across billions of pixels to produce clinically meaningful outputs. Building on this foundation, PathGen\cite{sun2024pathgen} shifted the focus to patch level semantics by generating fine grained descriptions for millions of tissue regions to capture nuanced morphological patterns, and HistGen \cite{guo2024histgen} further scaled patch level captioning across diverse datasets to improve descriptive accuracy and granularity. Meanwhile, unified generative frameworks such as CONCH \cite{lu2024visual}, Virchow \cite{vorontsov2024foundation} and CHIEF \cite{wang2024pathology} expanded the scope of multimodal pathology modeling by jointly learning to classify diagnoses, localize regions of interest and generate both slide level and patch level narratives within a single architecture. Most recently, GigaPath\cite{xu2024whole} and mSTAR\cite{xu2024multimodal} have combined extensive WSI collections with multimodal reasoning objectives to achieve robust generalizable performance across a wide range of tissue types and clinical tasks.

\subsection{Pixel-Level Visual Reasoning Models}
Most large language models (LLMs) are primarily designed to generate global textual descriptions and often lack the capacity for fine-grained, pixel-level visual reasoning \cite{xiao2024florence, ma2024groma}. To address this limitation, recent studies have introduced pixel-level supervision and enhanced vision-language alignment mechanisms to improve spatial reasoning capabilities \cite{zhang2025omg}. For example, LISA \cite{lai2024lisa} was one of the first models to enable pixel-level visual reasoning, although it is constrained to single-object understanding.
Subsequent models have extended these capabilities to handle more complex scenarios involving multiple instances. PixelLM \cite{ren2024pixellm}, GlaMM \cite{rasheed2024glamm}, PerceptionGPT \cite{pi2024perceptiongpt}, and GSVA \cite{xia2024gsva} support multi-object reasoning through improved multimodal alignment and spatial localization.
Beyond 2D reasoning, emerging models such as ZSVG3D \cite{yuan2024visual} and PRIMA \cite{wahed2024prima} explore 3D spatial understanding within multimodal frameworks. In parallel, models like OMG-LLaVA \cite{zhang2025omg} and GeoPixel \cite{shabbir2025geopixel} continue to advance pixel-level grounded visual reasoning, pushing the boundaries of interpretability in vision-language models.

Despite their potential for medical imaging, no existing work applies pixel-level visual reasoning to pathology. To bridge this gap, we propose the first multimodal visual reasoning model for pathological images, enhancing both interpretability and diagnostic reliability.

\definecolor{myblue}{RGB}{0, 138, 218}
\definecolor{myorange}{RGB}{204,85,0}
\section{Dataset Construction}

The advancement of visual reasoning tasks has led to the development of numerous pixel-level reasoning datasets in natural scenes, such as MUSE \cite{ren2024pixellm}, MMR \cite{jangmmr}, and ReasonSeg \cite{lai2024lisa}. However, extending pixel-level reaoning models to cancer diagnosis remains highly challenging, primarily due to the difficulty of obtaining suitable datasets. An ideal dataset must meet the following key criteria:
\noindent(1) \textit{Comprehensive cancer diagnostic annotations}, including class labels for cancer staging or subtype classification; 
\noindent(2) \textit{A large-scale collection of pathological image patches} to support the training of visual reasoning models;  
\noindent(3) \textit{Pixel-level annotations for muclei segmentation and classification}, enabling fine-grained, cell-level pathological analysis;  
\noindent(4) \textit{Expert-annotated textual question-answer pairs} that align with nuclear annotations and provide diagnostic reasoning in a structured format.

To address these challenges and advance visual reasoning in pathology, we introduce the \textbf{G}astric \textbf{A}denocarcinoma \textbf{D}iagnosis \textbf{V}isual \textbf{R}easoning (GADVR) dataset. Unlike existing cancer diagnosis datasets that primarily provide WSI-level or patient-level labels, GADVR incorporates both pixel-level nuclei annotations and patch-level text annotations, enabling the development of cell-level multimodal reasoning models for fine-grained interpretability analysis.

\subsection{Cancer Diagnosis Data}

Existing multimodal cancer diagnosis datasets primarily provide patch-level annotations but often lack nuclear segmentation and classification labels. Conversely, datasets designed for nuclear segmentation and classification typically contain only a limited number of image patches. For example, PanNuke \cite{gamper2019pannuke} consists of only approximately 7,000 patches, each with a resolution of $256 \times 256$ pixels. These limitations hinder the development of large-scale visual reasoning models.

To address these challenges, we construct the GADVR dataset based on PatchGastricADC22 \cite{tsuneki2022inference}, a dataset for gastric adenocarcinoma subtype diagnosis. PatchGastricADC22 consists of 262,777 pathological image patches, each with a resolution of $600 \times 600$ pixels at $\times$40 magnification. These patches are categorized into nine gastric adenocarcinoma subtypes based on their corresponding WSI labels, including well-differentiated, moderately differentiated tubular adenocarcinoma, and papillary adenocarcinoma, among others. The large dataset size and comprehensive subtype annotations provide a strong foundation for developing multimodal reasoning models. Additionally, PatchGastricADC22 includes WSI-level diagnostic reports, which serve as a valuable resource for generating text-based question-answer pairs.

\begin{figure}[t!]
    \centering
    \includegraphics[width=3.5in]{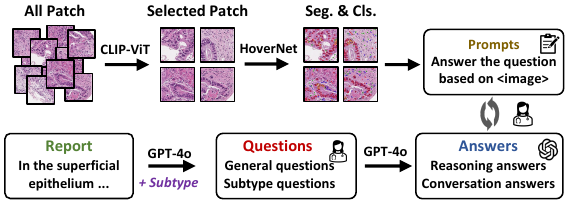}
    \caption{\textbf{Overview of the GADVR dataset construction process.} The dataset is built by selecting representative patches using pre-trained CLIP-ViT \cite{sun2024pathgen}, followed by nuclei segmentation and classification with HoverNet \cite{graham2019hover}. Text reports are processed with GPT-4o to generate cell-level QA pairs. ``Seg.” and ``Cls.” denote segmentation and classification, respectively.}
    \vspace{-0.4cm}
    \label{fig2}
\end{figure}

\subsection{GADVR Generation Pipeline}

Our dataset construction process includes: 1) pixel-level nuclear segmentation, classification label generation, and 2) cell-level text question-answer (QA) pair generation. The specific process is shown in Fig. \ref{fig2}.

\begin{figure*}[t!]
    \centering
    \includegraphics[width=7.1in]{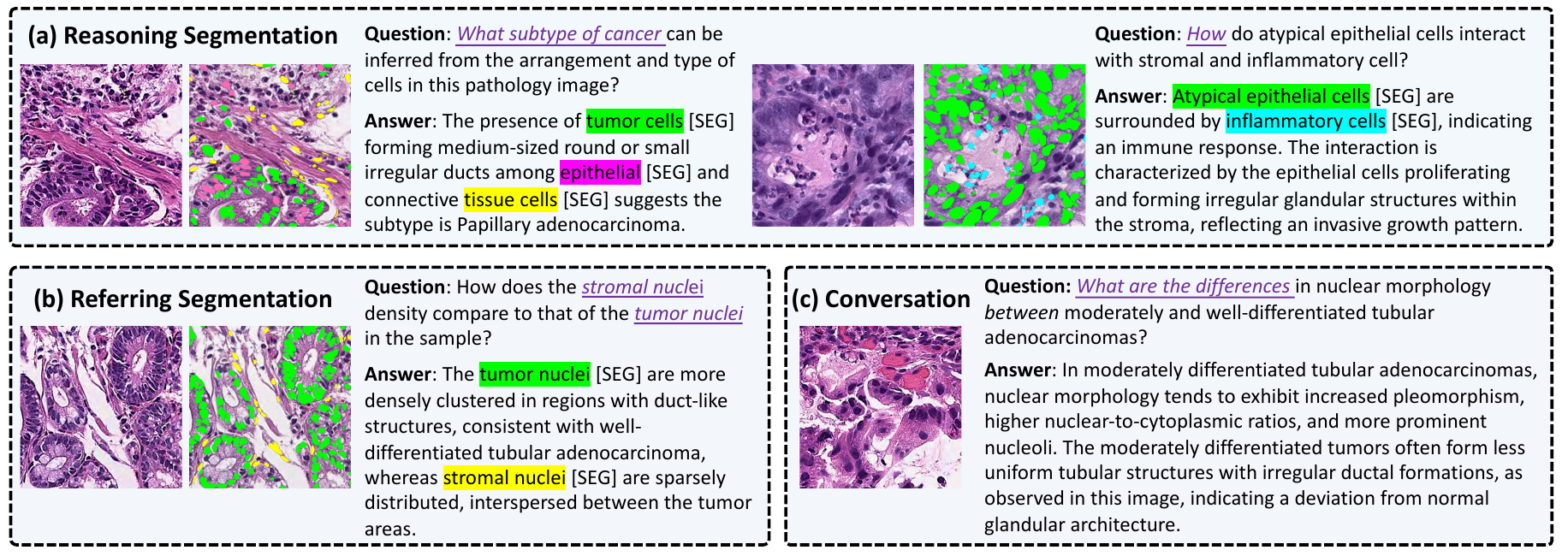}
    \caption{\textbf{The examples from our constructed GADVR dataset:} (a) showcases two reasoning-based segmentation examples, (b) presents a referring segmentation example, and (c) illustrates a conversation-based example.}
    \vspace{-0.3cm}
    \label{fig3}
\end{figure*}

\subsubsection{Nuclei Annotation Generation}

Since patch-level labels are inherited from WSI annotations, certain patches may exhibit weak correlations between the local tissue texture and the assigned subtype label. To address this, we first apply a pretrained CLIP-ViT model \cite{sun2024pathgen}, leveraging the corresponding original diagnostic reports to identify and remove weakly correlated patches. The remaining high-correlation patches are then used to generate pixel-level nuclei segmentation and classification labels using a pretrained HoverNet model \cite{graham2019hover}. To determine the most suitable pretraining dataset for HoverNet, we evaluate models trained on several publicly available datasets, including ConSeP, Lizard, and PanNuke. Based on expert pathological review, PanNuke is selected for pretraining due to its superior segmentation performance.

To further improve nuclei classification accuracy, two pathologists manually correct misclassified nuclei in approximately 200 patches per category. These corrections are integrated into a human-in-the-loop refinement process to optimize the classification labels. Subsequently, three independent pathologists assess the final nuclei segmentation and classification results, confirming that the constructed dataset achieved excellent quality. A detailed quality assessment is provided in the subsection Quality Control.

\subsubsection{Question-Answer Pair Generation}
To enable fine-grained interpretability, our dataset incorporates cell-level question-answer (QA) pairs designed to support pixel-wise pathological reasoning. The data generation process is illustrated in Fig. \ref{fig2}.  

\noindent\textbf{Question Generation}
We employ GPT-4o to generate a diverse set of diagnostic questions based on pathology reports and specific cancer subtypes. The questions are categorized into two types: \textbf{(1) General questions}, which focus on describing pathological image features such as nuclear morphology, spatial distribution, and tissue architecture; and \textbf{(2) Subtype-specific questions}, which target diagnostic reasoning, such as inferring the cancer subtype from nuclear morphology and distribution patterns. These two categories are complementary and progressive, where understanding general morphological characteristics provides a foundation for accurate subtype diagnosis. In total, approximately 200 questions are generated.

\noindent\textbf{Answer Generation}
For each image, three questions are randomly sampled from the question pool and answered using GPT-4o. To improve clinical relevance and contextual accuracy, both the selected questions and the corresponding nuclei segmentation masks are provided as inputs to the model. The prompt design is iteratively refined to guide the model in generating answers that explicitly incorporate nuclear categories and spatial distribution patterns. This ensures that the generated responses are closely aligned with the visual information conveyed by the segmentation masks.

\subsection{Dataset Presentation}

Fig. \ref{fig3} illustrates representative examples from the GADVR dataset, which integrates cell-level image-text annotations to support interpretable diagnosis of gastric adenocarcinoma. The dataset is designed to facilitate three core tasks: reasoning segmentation, referring segmentation, and conversation task.
In the reasoning segmentation, the model aligns pixel-level visual predictions with textual answers, enabling explicit visual reasoning. The model not only localizes cells, but also associates them with corresponding textual descriptions that explain their diagnostic significance.
The referring segmentation focuses on identifying specific nuclei types based on given textual references.
Finally, the conversation task enhances interpretability by simulating a dialogue grounded in pathology knowledge.

\begin{figure}[t!]
    \centering
    \includegraphics[width=3.6in]{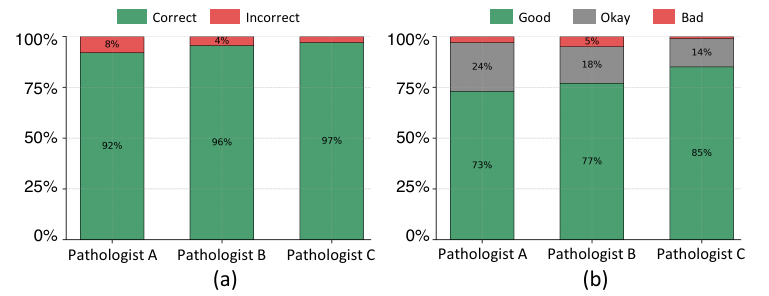}
    \caption{\textbf{Summary of pathologist assessments.} (a) Segmentation correctness with proportions judged correct and incorrect. (b) Quality ratings with distribution of Good, Okay, and Bad.}
    \vspace{-0.4cm}
    \label{fig11}
\end{figure}

\begin{figure*}[t!]
    \centering
    \includegraphics[width=7.2in]{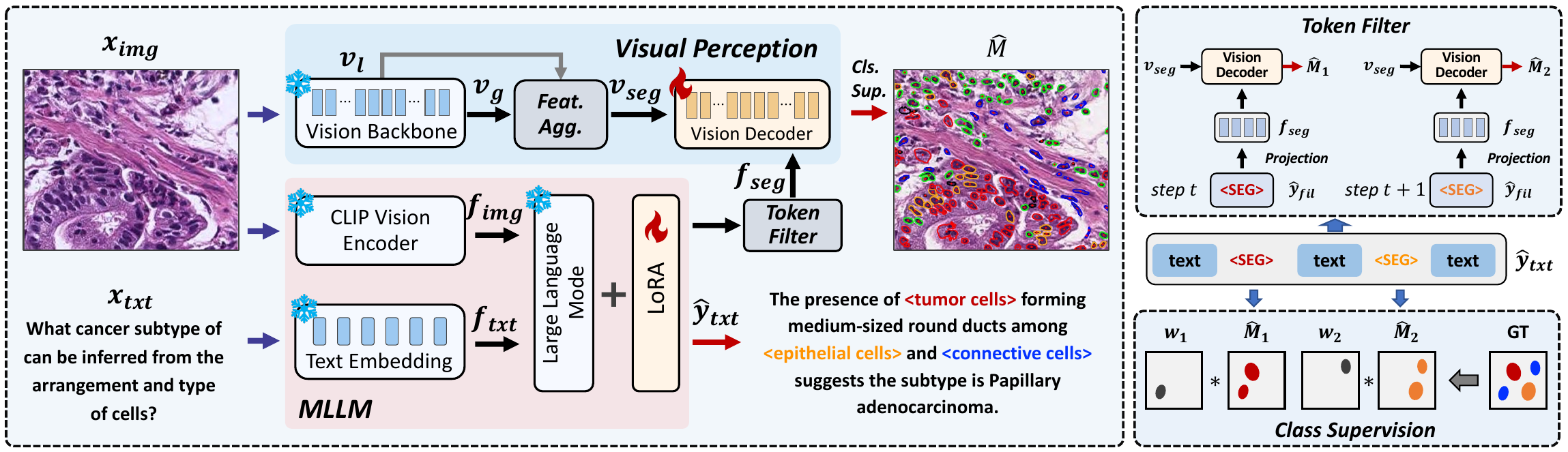}
    \caption{\textbf{Overview of the proposed framework.} The model combines SAM for segmentation and a ViT-based MLLM for multimodal reasoning, enabling text-guided segmentation through visual-text interaction.}
    \vspace{-0.2cm}
    \label{fig4}
\end{figure*}

\subsection{Quality Control}


To assess the quality of the constructed GADVR dataset, we invited three independent external pathologists to evaluate a randomly selected subset of 200 image samples. Each sample includes nuclei segmentation masks and their corresponding QA pairs. For each sample, the pathologists assessed the overall image quality (good, okay, or bad) and verified whether the key structures were correctly segmented and free of major artifacts or errors. In addition, the quality of QA pairs was evaluated with respect to their clinical meaningflness, image dependency and correctness.
Fig. \ref{fig11} summarizes the evaluation results. The overall distribution remains consistent across the three experts. Specifically, over 92\% of the segmentation masks are rated as Correct, and more than 95\% of the QA pairs are judged to be of Okay or Good quality.
These results demonstrate that the dataset maintains high quality in both segmentation accuracy and textual reasoning, making it a reliable resource for interpretable pathology model development.

\section{Methods}
Although multimodal large language models (LLMs) can enhance feature extraction through vision-language interaction, existing methods typically support only unimodal text output and lack the capability for pixel-level reasoning. To address this limitation, we introduce an visual perception module on top of the existing LLM framework, enabling mask prediction for images. In detail, the architecture of our proposed model is illustrated in Fig. \ref{fig4}.
\subsection{Visual Perception Module}

\begin{table*}[t]
    \centering
    \small
    \renewcommand{\arraystretch}{1.15} 
    \setlength{\tabcolsep}{4.0pt} 
    \caption{\textbf{The reasoning segmentation performance comparison on the GADVR benchmark.} ``E2E" stands for end-to-end training. $^{\dag}$Ov-Seg is a two-stage approach; it first predicts categories, which are then used as prompts for the segmentation. TFLPOs are tested based on $600 \times 600$ input.}
    \begin{tabular}{l|c|cc|cc|cc|cc|cc|cc|c}
        \toprule
        \multirow{2}{*}{Method} & \multirow{2}{*}{TFLOPs} & \multicolumn{2}{c|}{Validation} & \multicolumn{2}{c|}{Test Overall} & \multicolumn{2}{c|}{Neoplastic} & \multicolumn{2}{c|}{Inflammatory} & \multicolumn{2}{c|}{Connective} & \multicolumn{2}{c|}{Epithelial} & \multirow{2}{*}{Publication} \\
        \cmidrule(lr){3-6} \cmidrule(lr){7-14}
        & & gIoU & cIoU & gIoU & cIoU & gIoU & cIoU & gIoU & cIoU & gIoU & cIoU & gIoU & cIoU \\
        \midrule
        Ov-Seg$^{\dag}$\cite{liang2023open}   & --   & 0.356   & 0.505    & 0.383    & 0.515  & 0.468 & 0.522 & 0.365 & 0.344 & 0.101 & 0.132 & 0.069 & 0.121 & CVPR'23  \\
        LISA-7B \cite{lai2024lisa}    & 11.61  & 0.402    & 0.520    & 0.429    & 0.549  & 0.543 & 0.574 & 0.392 & 0.420 & 0.183 & 0.221 & 0.080 & 0.113 & CVPR'24  \\
        PixelLM \cite{ren2024pixellm} & 15.29  & 0.411 & 0.530 & 0.438 & 0.558  & 0.550 & 0.584 & 0.400 & 0.429 & 0.190 & 0.237 & 0.098 & 0.145 & CVPR'24  \\
        GSVA \cite{xia2024gsva}       & 15.61   & 0.367    & 0.516    & 0.409   & 0.520  & 0.509 & 0.543 & 0.355 & 0.363 & 0.104 & 0.187 & 0.054 & 0.096 & CVPR'24  \\
        MMR-7B \cite{jangmmr}     & 11.14   & \underline{0.587} & \underline{0.698} & \underline{0.602} & \underline{0.714} & \underline{0.678} & \underline{0.710} & \textbf{0.536} & \textbf{0.548} & \underline{0.340} & \underline{0.384} & \underline{0.200} & \underline{0.264} & ICLR'25  \\
        \rowcolor{gray!25} 
        PathMR-7B  & 11.14   & \textbf{0.603}  & \textbf{0.715}  & \textbf{0.629}  &  \textbf{0.734} & \textbf{0.689} & \textbf{0.722} & \underline{0.529} & \underline{0.533} & \textbf{0.351} & \textbf{0.402} & \textbf{0.211} & \textbf{0.287} &  Ours      \\
        \midrule
        LISA-13B \cite{lai2024lisa}   & 16.84  & 0.409 & 0.529 & 0.436  & 0.559 & 0.554 & 0.583 & 0.394 & 0.422 & 0.183 & 0.221 & 0.075 & 0.101 & CVPR'24  \\
        MMR-13B \cite{jangmmr} & 17.21 &  \underline{0.598}   &  \underline{0.708}    & \underline{0.613}   & \underline{0.724}  & \textbf{0.709} & \textbf{0.739} & \underline{0.571} & \underline{0.581} & \underline{0.400} & \underline{0.447} & \underline{0.232} & \underline{0.328} & ICLR'25  \\
        \rowcolor{gray!25} 
        PathMR-13B  & 17.21 &
        \textbf{0.624}  & \textbf{0.723}  & \textbf{0.633}  &  \textbf{0.740}  & \underline{0.694} & \underline{0.737} & \textbf{0.685} & \textbf{0.726} & \textbf{0.401} & \textbf{0.467} & \textbf{0.389} & \textbf{0.440} & Ours  \\
        \bottomrule
    \end{tabular}
    \vspace{-0.2cm}
    \label{tab:con1}
\end{table*}

\begin{figure*}[t!]
    \centering
    \includegraphics[width=7.1in]{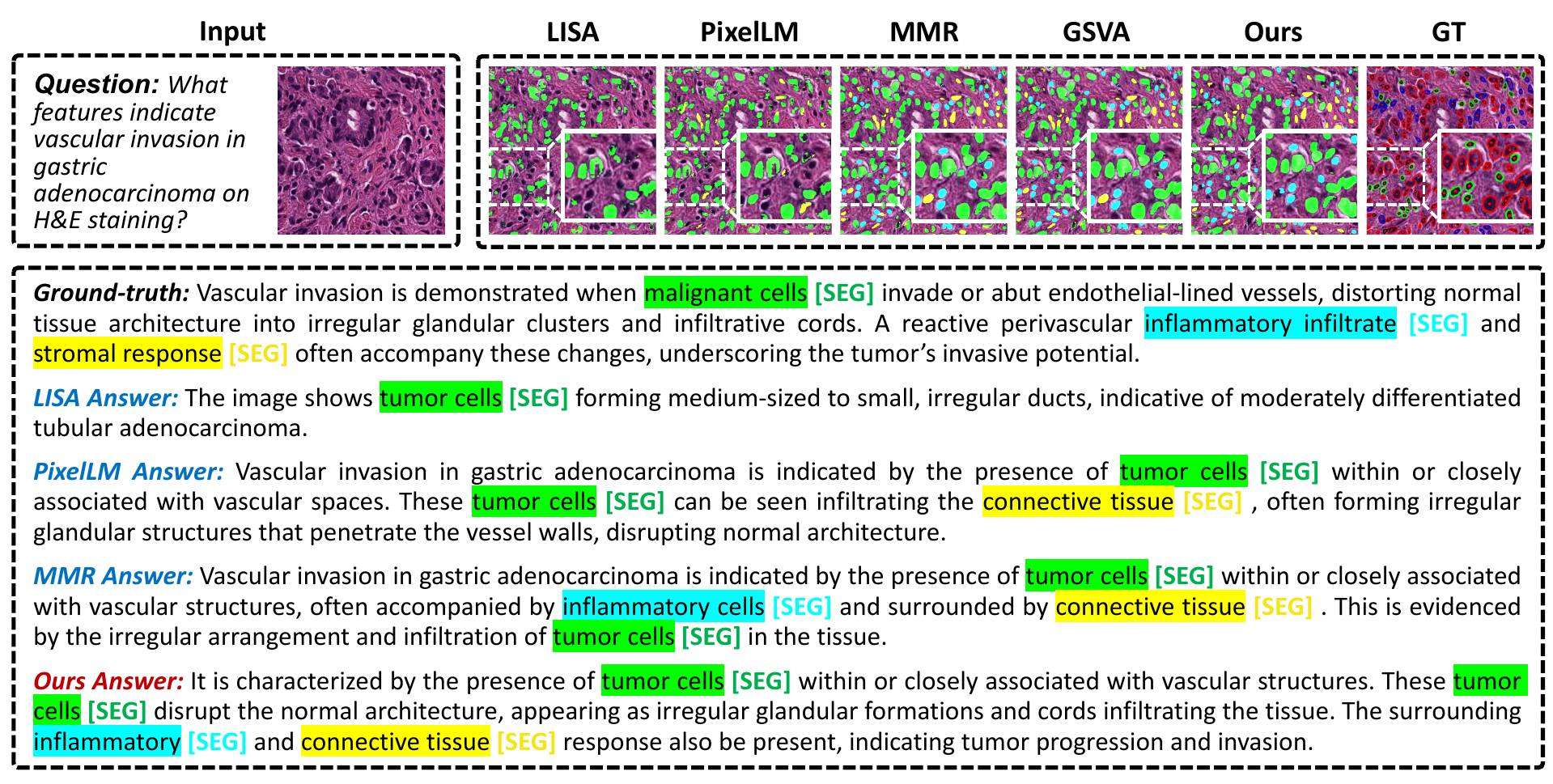}
    \caption{\textbf{Qualitative comparison of reasoning segmentation results across different models.} The input includes an H\&E-stained pathology image paired with a corresponding question. The output consists of a generated textual response accompanied by aligned masks. In the GT, the \textcolor{red}{red}, \textcolor{blue}{blue}, and \textcolor{green}{green contours} represent \textcolor{red}{tumor}, \textcolor{blue}{connective}, and \textcolor{green}{inflammatory cells}, respectively.}
    \vspace{-0.4cm}
    \label{fig5}
\end{figure*}

\noindent\textbf{Vision Backone}
The visual perception module employs a vision backbone to extract deep image features, as illustrated in the blue box in Fig. \ref{fig4}. We utilize the SAM vision encoder to capture multi-scale representations from the input image \( x_{img} \in \mathbb{R}^{h \times w \times 3} \). To effectively aggregate both global and local information, we incorporate the feature aggregation layers proposed in MMR \cite{jangmmr}. Specifically, the global feature representation is denoted as \( v_g \in \mathbb{R}^{h/16 \times w/16 \times c} \), while the early-stage local features are represented as \( v_l \in \mathbb{R}^{h/16 \times w/16 \times c'} \), where \( h \) and \( w \) are the height and width of the input image, and \( c \) and \( c' \) denote the channel dimensions. The final segmentation feature representation \( v_{seg} \), is computed as follows:  
\begin{equation}
    v_{seg} = v_g + \text{Conv}(v_l),
\end{equation}
where \(\text{Conv}(\cdot)\) denotes convolutional operations applied to the local feature representation. 

\noindent\textbf{Vision Decoder} 
The vision decoder takes as input both the segmentation feature \( v_{seg} \) from the vision backbone and the feature representation \( f_{seg} \) obtained from the MLLM module to generate the final segmentation mask:
\begin{equation}
    \hat{M} = \text{Dec}(\text{Concat}(v_{seg}, f_{seg})),
\end{equation}
where \(\text{Dec}(\cdot)\) represents the decoder function, and \(\text{Concat}(\cdot)\) denotes the concatenation operation. 

\noindent\textbf{Classification Supervision}
In visual reasoning tasks, pixel-level segmentation results often contain significant prediction noise, including isolated prediction masks and irregular nuclei morphologies. To address the isolated segmentation results caused by uncertainty in category predictions, we introduce a classification supervision method before calculating the segmentation loss, as shown in Fig. \ref{fig4} (b). 

In the design, the text output \(\hat{y}_{txt}\) undergoes token filtering to extract the representations corresponding to the $<seg>$ token, yielding \(\hat{y}_{fil}\). This filtered representation is then mapped to the MLLM output \(f_{\text{seg}}\), which serves as input for the visual decoder. However, during this process, the network incrementally predicts the $<seg>$ token and outputs corresponding binary segmentation masks \(\hat{M}_i\). This sequential prediction leads to a lack of awareness of subsequent categories, causing the model to classify other categories as the current one incorrectly. To mitigate this issue, we apply classification supervision by introducing additional penalties for incorrect predictions when the model mistakenly identifies other categories as the current class. This ensures greater accuracy in category predictions. The segmentation loss in this process is defined as:
\begin{equation}
    \mathcal{L}_{\text{mask}} = \lambda_1 \text{BCE}(\hat{M}, M, W) + \lambda_2 \text{DICE}(\hat{M}, M, W),
\end{equation}
where \(\text{BCE}(\cdot)\) and \(\text{DICE}(\cdot)\) represent the binary cross-entropy and Dice loss functions, respectively. \(W\) represents the penalty weight set to 1.5, and the hyperparameters \(\lambda_1\) and \(\lambda_2\) are set to 2 and 0.5 in the paper.

\noindent\textbf{Consistency Constraint}
To enhance the completeness of nuclear morphology, we introduce a consistency constraint that enforces adjacent pixels to have similar category predictions. In this step, for each pixel, we evaluate whether its predicted category is consistent with that of its four immediate neighbors (up, down, left, and right). This approach aims to smooth the prediction map, thereby reducing local noise and discontinuities. Specifically, let \( p_{ij} \) denote the prediction probability of the pixel at position \((i,j)\). The consistency loss is defined as:
\begin{equation}
\begin{split}
L_{\text{con}} = \frac{1}{N} \sum_{i,j} \Bigl( & \, \left| p_{i,j} - p_{i+1,j} \right| + \left| p_{i,j} - p_{i-1,j} \right| \\
& + \left| p_{i,j} - p_{i,j+1} \right| + \left| p_{i,j} - p_{i,j-1} \right| \Bigr)
\end{split}
\end{equation}
where \(N\) represents the total number of valid neighboring pixel pairs. The objective of this loss is to minimize the differences in prediction probabilities between neighboring pixels, thus ensuring a more consistent class distribution across the predicted map.

\begin{table*}[t]
    \centering
    \small
    \renewcommand{\arraystretch}{1.15} 
    \setlength{\tabcolsep}{8pt} 
    \caption{\textbf{The reasoning segmentation performance comparison on PathGen benchmark.}}
    \begin{tabular}{l|cc|cc|cc|cc|cc}
        \toprule
        \multirow{2}{*}{Method} & \multicolumn{2}{c|}{Neoplastic} & \multicolumn{2}{c|}{Inflammatory} & \multicolumn{2}{c|}{Connective} & \multicolumn{2}{c|}{Epithelial} & \multicolumn{2}{c}{Overall} \\
        \cmidrule(lr){2-3} \cmidrule(lr){4-5} \cmidrule(lr){6-7} \cmidrule(lr){8-9} \cmidrule(lr){10-11}
        & gIoU & cIoU & gIoU & cIoU & gIoU & cIoU & gIoU & cIoU & gIoU & cIoU \\
        \midrule
        LISA-LLaVA-13B \cite{lai2024lisa}  & 0.517 & 0.584 & 0.477 & 0.550 & 0.429 & 0.507 & 0.233 & 0.318 & 0.484 & 0.519 \\
        MMR-LLaVA-13B$^{\ddag}$ \cite{jangmmr} &  0.573 & 0.645 & 0.521 & 0.607 & 0.483 & 0.577 & 0.304& 0.397 & 0.545 & 0.626 \\
        \rowcolor{gray!25} 
        PathMR-LLaVA-13B (ours) & 
        0.596 & 0.674 & 0.537 & 0.621 & 0.498 & 0.597 & 0.334 & 0.421 &  0.560 & 0.635 \\
        \rowcolor{gray!25} 
        PathMR-Qwen (ours) & 0.612 & 0.681 & 0.551 & 0.637 & 0.511 & 0.607 & 0.344 & 0.434 & 0.579 & 0.641 \\
        \bottomrule
    \end{tabular}
    \vspace{-0.2cm}
    \label{tab:con2}
\end{table*}

\begin{figure*}[h!]
    \centering
    \includegraphics[width=6.6in]{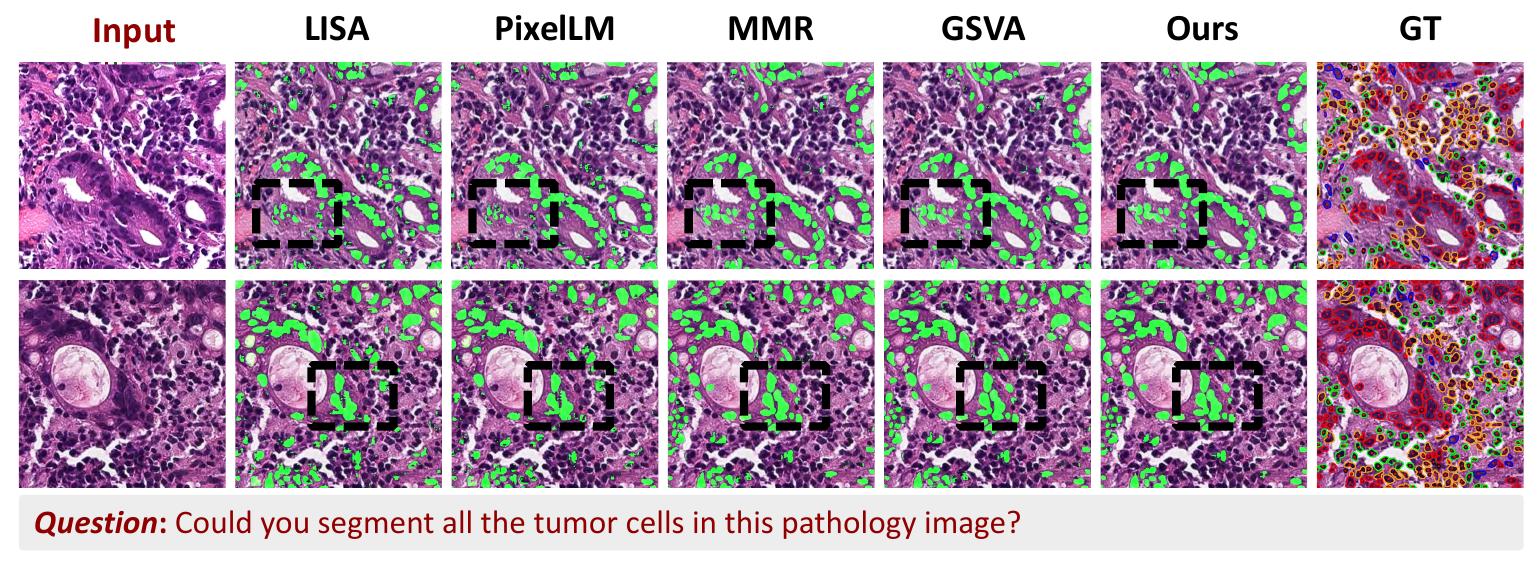}
    \caption{\textbf{Qualitative comparison of visual reasoning results across different models.} The red text represents the model's input, including the image and question. The black text corresponds to the segmentation prediction and the ground truth.}
    \vspace{-0.4cm}
    \label{fig7}
\end{figure*}

\subsection{Multimodal Large Language Model}
To integrate textual and visual modalities, the input image \( x_{img} \) is first processed by a CLIP vision encoder and subsequently mapped into the LLM feature space, producing visual token embeddings \( f_{img} \in \mathbb{R}^{N_{img} \times d} \). For simplicity, the mapping layer is omitted in Fig. \ref{fig4}. Meanwhile, the textual input \( x_{txt} \) is tokenized and embedded using the language model’s tokenizer, yielding text token embeddings \( f_{txt} \in \mathbb{R}^{N_{txt} \times d} \). Then, the visual and textual embeddings are then concatenated and fed into the large language model \( F \), which generates the textual response:  
\begin{equation}
    \hat{y}_{txt} = F(Concat(f_{img}, f_{txt})).
\end{equation}
Notably, \(\hat{y}_{txt} \) includes both the generated textual response and specialized [SEG] tokens that correspond to segmented objects. Based on this output, the textual loss \( L_{txt} \) is computed as follows:
\begin{equation}
    \mathcal{L}_{txt} = CE(\hat{y}_{txt}, y_{txt}).
\end{equation}
 
Finally, the model is optimized end-to-end using a combined objective function that integrates both loss in text generation \(\mathcal{L}_{\text{txt}}\) and loss in consistency constraint \(\mathcal{L}_{\text{con}}\):
\begin{equation}
    \mathcal{L} = \lambda_{\text{mask}} \mathcal{L}_{mask} +\lambda_{\text{txt}} \mathcal{L}_{txt} + \lambda_{\text{con}} \mathcal{L}_{con}, 
\end{equation}
where the coefficients \(\lambda_{\text{mask}}\), \(\lambda_{\text{txt}}\) and \(\lambda_{\text{con}}\) are used to balance the losses. By default, these weights are set to 1.

\section{Experiments}

\subsection{Datasets}

We evaluate the performance of our model on two datasets: the proposed GADVR dataset and the public PathGen dataset.
The GADVR\footnote{The GADVR dataset is publicly available at: \url{https://huggingface.co/datasets/zhangye-zoe/GADVR}} contains approximately 190,000 pathology image patches and over 547,000 image-text pairs, all derived from WSIs of gastric adenocarcinoma. To ensure consistent evaluation, we split the dataset by WSI into training, validation, and test subsets following an 8:1:1 ratio. Specifically, the training set contains 149,928 patches and 447,170 image-text pairs, the validation set contains 14,814 patches and 44,080 pairs, and the test set includes 18,754 patches and 56,067 pairs.

To evaluate the generalizability of our model across diverse tissue types, we further conduct training using the PathGen dataset \cite{sun2024pathgen}, a pan-cancer resource that includes pathology images from over 25 different organs. We randomly sample 50,000 images from PathGen to form an independent training set. Since PathGen does not provide cell-level segmentation annotations, we generate segmentation labels using the same pipeline employed for GADVR. This allows our model to be trained and evaluated consistently across both datasets in terms of cell-level segmentation and image-text reasoning.

\begin{table*}[h]
    \centering
    \small
    \renewcommand{\arraystretch}{1.15} 
    \setlength{\tabcolsep}{6pt} 
    \caption{\textbf{The referring segmentation performance comparison on GADVR dataset.}}
    \begin{tabular}{l|cc|cc|cc|cc|cc}
        \hline
        \multirow{2}{*}{Methods} & \multicolumn{2}{c|}{Neoplastic} & \multicolumn{2}{c|}{Inflammatory} & \multicolumn{2}{c|}{Connective} & \multicolumn{2}{c|}{Epithelial} & \multicolumn{2}{c}{Overall} \\
        \cline{2-11}
        & gIoU & cIoU & gIoU & cIoU & gIoU & cIoU & gIoU & cIoU & gIoU & cIoU \\
        \hline
        OV-Seg$^{\dag}$ \cite{liang2023open}   & 0.508   & 0.564    & 0.048    & 0.036    & 0.076    & 0.062    & 0.062    & 0.058    & 0.253    & 0.405    \\
        LISA-7B \cite{lai2024lisa}    & 0.529 & 0.561 & 0.383   & 0.402    & 0.180    & 0.210    & 0.008    & 0.113    & 0.337    & 0.503    \\
        PixelLM \cite{ren2024pixellm} & 0.538   & 0.571    & 0.390    & 0.408    & 0.186    & 0.224    & 0.095    & 0.139    & 0.346    & 0.511    \\
        GSVA \cite{xia2024gsva} & 0.494    & 0.551   & 0.127    & 0.209    & 0.093   & 0.072    & 0.025    & 0.089    & 0.288   & 0.454    \\
        MMR-7B$^{\ddag}$ \cite{ren2024pixellm}  & \underline{0.655}    & \underline{0.736}    & \underline{0.474}    & \underline{0.503}    & \underline{0.474}    & \underline{0.498}    & \underline{0.240}    & \underline{0.300}    & \underline{0.527}    & \underline{0.634}    \\
        PathMR-7B    & \textbf{0.672}    & \textbf{0.756}    & \textbf{0.491}    & \textbf{0.513}    & \textbf{0.497}   & \textbf{0.522}    & \textbf{0.264}    & \textbf{0.314}    & \textbf{0.539}    & \textbf{0.657}    \\
        \midrule
        LISA-13B \cite{lai2024lisa}   & 0.540   & 0.571    & 0.386    & 0.412    & 0.187    & 0.219    & 0.076    & 0.113    & 0.341    & 0.517    \\
        MMR-13B$^{\ddag}$ \cite{ren2024pixellm} & \underline{0.695} & \underline{0.724} & \underline{0.551}  & \underline{0.561}  & \underline{0.394}   & \underline{0.420}   &  \underline{0.228}    & \underline{0.290}   & \underline{0.515} & \underline{0.664}    \\
        BioMedParse \cite{zhao2024foundation} & 0.445    & 0.478    & 0.260   & 0.283    & 0.214    & 0.232    & 0.161   & 0.177  & 0.254 & 0.461    \\
        PathMR-13B  & \textbf{0.711}    & \textbf{0.746}    & \textbf{0.571}  & \textbf{0.589}    & \textbf{0.402}   & \textbf{0.431}    & \textbf{0.238}    & \textbf{0.311}    & \textbf{0.530}    & \textbf{0.682}    \\
        \hline
    \end{tabular}
    \vspace{-0.2cm}
    \label{tab:con3}
\end{table*}

\subsection{Implementation Details}
For the visual perception module, we employ ViT-H SAM \cite{kirillov2023segment} as both the visual encoder and decoder. Within the multimodal LLM framework, we use a pre-trained CLIP-ViT-L/14-336 model \cite{radford2021learning} as the vision encoder, while LLaVA-v0.5-7B and llava-llama-2-13B-chat-lightning-preview \cite{liu2024visual} serve as the language models. To improve fine-tuning efficiency, we integrate LoRA. 
Following the training setup of LISA \cite{lai2024lisa}, we use the AdamW optimizer with a learning rate of 0.0003. For the 7B model, we set the batch size to 16 with 1200 steps per epoch, while for the 13B model, we use a batch size of 6 with 3125 steps per epoch. The LoRA rank is fixed at 8. Training is conducted on 8 A100 GPUs, with a total training time of approximately 1 day for the 7B model and 1.5 days for the 13B model.

\subsection{Baselines and Evaluation Metrics}

\noindent\textbf{Baselines}
To thoroughly assess our model's performance across reasoning segmentation, referring segmentation, and dialogue tasks, we benchmark it against state-of-the-art visual reasoning models designed for natural image domains, including Ov-Seg \cite{liang2023open}, LISA \cite{lai2024lisa}, PixelLM \cite{ren2024pixellm}, GSVA \cite{xia2024gsva}, and MMR \cite{jangmmr}, all of which support these three tasks. Additionally, to enable a comprehensive comparison with models tailored for pathology image analysis, we incorporate the multimodal LLMs BiomedParse \cite{zhao2024foundation} and LLaVA-Med \cite{li2023llava} as additional baselines for the referring segmentation and conversation tasks. To ensure a fair comparison, we fine-tune all visual reasoning models on our dataset. 

\noindent\textbf{Evaluation Metrics} 
We evaluate our model's performance on both segmentation and text generation tasks. For segmentation, we use gIoU and cIoU as evaluation metrics, following prior works \cite{ren2024pixellm, jangmmr}. For text conversation, we assess performance using BLEU-4 and F1 scores. Throughout the paper, we denote the best performing result in \textbf{ bold} and the second-best result with \underline{underline}.

\begin{table}[t]
    \centering
    \small
    \renewcommand{\arraystretch}{1.2} 
    \setlength{\tabcolsep}{6pt} 
    \begin{tabular}{l|cc|cc}
        \hline
        \multirow{2}{*}{Methods} & \multicolumn{2}{c|}{Validation} & \multicolumn{2}{c}{Test} \\
        \cline{2-5}
        & BLEU-4 & F1 & BLEU-4 & F1 \\
        \hline
        LISA-7B \cite{lai2024lisa} & 0.276 & 0.575 & 0.278 & 0.582 \\
        LISA-13B \cite{lai2024lisa} & 0.279 & \underline{0.586} & \underline{0.286} &\underline{0.591} \\
        PixelLM \cite{ren2024pixellm}  & 0.271 & 0.574 & 0.268 & 0.578   \\
        GSVA \cite{xia2024gsva} & 0.258 & 0.566 & 0.261 & 0.572 \\
        MMR-7B \cite{jangmmr} & 0.265 & 0.567 & 0.270 & 0.584 \\
        MMR-13B \cite{jangmmr} & 0.280 & 0.576 & 0.281 & 0.590 \\
        LLaVA-Med \cite{li2023llava} & \underline{0.284} & \underline{0.586} & 0.283 & \underline{0.591} \\
        \hline
        PathMR (ours) & \textbf{0.301} & \textbf{0.609} & \textbf{0.302} & \textbf{0.602} \\
        \hline
    \end{tabular}
    \caption{\textbf{The conversation performance comparison on GADVR dataset.}}
    \vspace{-0.3cm}
    \label{tab:con4}
\end{table}

\subsection{Comparative Experiments}
\subsubsection{Results on Reasoning Segmentation}

Table \ref{tab:con1} presents a comprehensive comparative analysis of our method against state-of-the-art visual reasoning approaches. In this experiment, we evaluate models at both the 7B and 13B parameter scales and benchmark their performance against existing methods. The results demonstrate that our approach consistently achieves superior performance across multiple evaluation metrics.
Furthermore, to assess the generalization capability of our method across diverse tissue types, we conducte additional experiments on the PathGen dataset, incorporating LLaVA and Qwen as the underlying large language models. As shown in Table \ref{tab:con2}, our approach maintains a notable advantage over LISA and MMR, achieving improvements of more than 1.5\% and 2.4\% in the gIoU metric, respectively.

Furthermore, Fig. \ref{fig5} presents a qualitative comparison of visual reasoning outputs across different models. The visualization demonstrates that our method achieves superior alignment between textual descriptions and segmentation results. Specifically, our model accurately identifies tumor cells, inflammatory cells, and connective tissue, ensuring consistency between the generated response and the segmentation mask. In contrast, other methods exhibit notable limitations: LISA and PixelLM fail to capture crucial regions, while GSVA produces less coherent segmentation. Our approach not only addresses these shortcomings but also provides a more comprehensive and precise interpretation.

\subsubsection{Results on Referring Segmentation}

To evaluate the effectiveness of our model in referring segmentation, we conduct comprehensive experiments on the GADVR benchmark, and report the results in Table~\ref{tab:con3}. 
Our proposed PathMR model achieves state-of-the-art performance across all categories on both 7B and 13B scales. Specifically, PathMR-13B attains the highest overall scores of 0.530 gIoU and 0.682 cIoU, surpassing the previous best model MMR-13B, which scores 0.515 and 0.664, respectively. In particular, PathMR demonstrates strong robustness on rare categories, such as epithelial cells, which many baseline methods including LISA and OV-Seg fail to segment accurately. 

In addition to the quantitative gains, qualitative comparisons in Fig.~\ref{fig7} further highlight PathMR's superior reasoning ability. Compared to existing methods, our model yields sharper, more complete, and category-aligned segmentation masks, particularly in regions with dense tumor cells or ambiguous boundaries. Unlike other approaches that tend to under-segment or generate noisy masks, PathMR produces precise predictions that better align with the ground truth.

\begin{figure}[t!]
    \centering
    \includegraphics[width=3.2in]{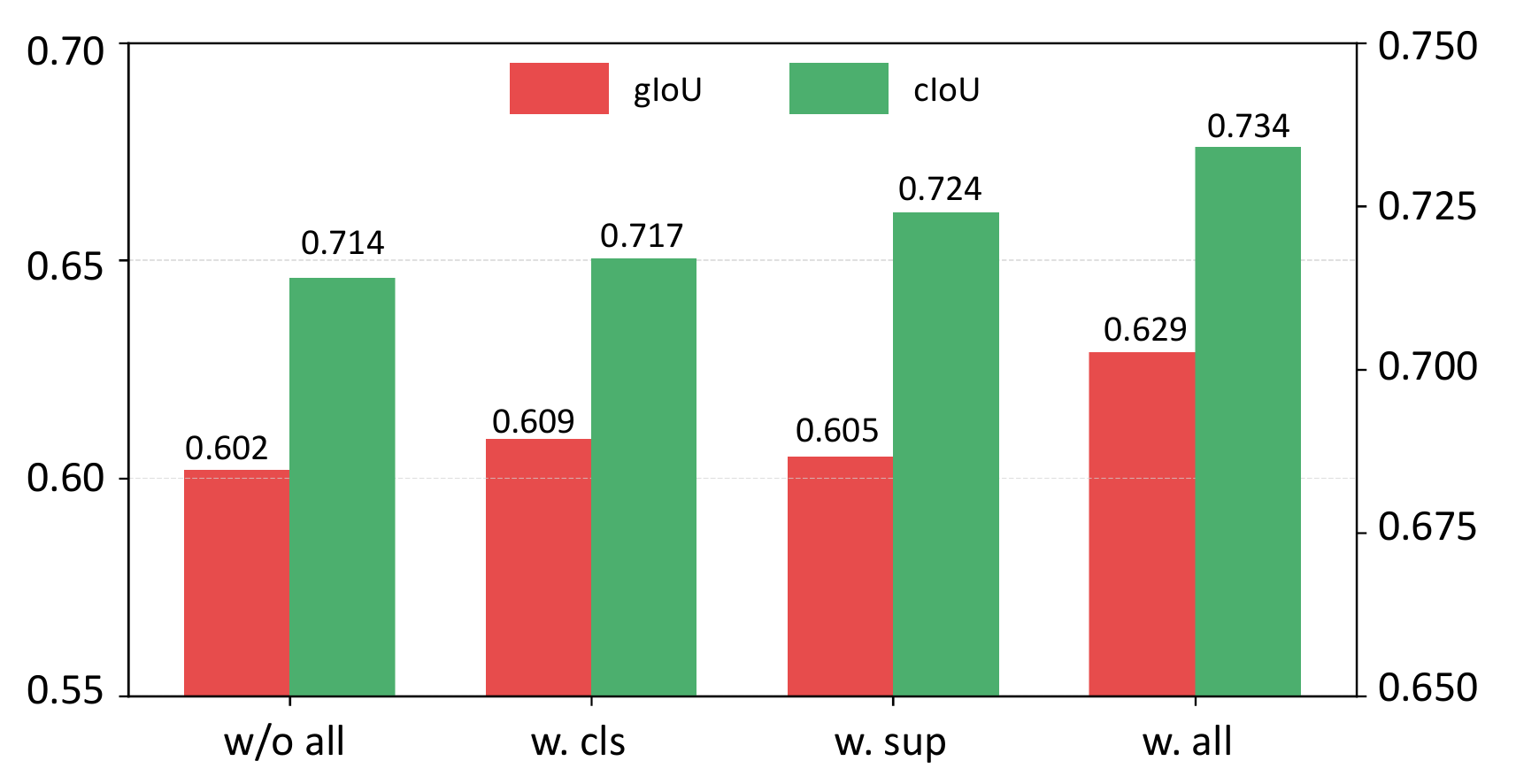}
    \caption{\textbf{Ablation study of module designs on reasoning segmentation on GADVR dataset.}}
    \vspace{-0.4cm}
    \label{fig8}
\end{figure}


 \begin{figure*}[t!]
    \centering
    \includegraphics[width=6.6in]{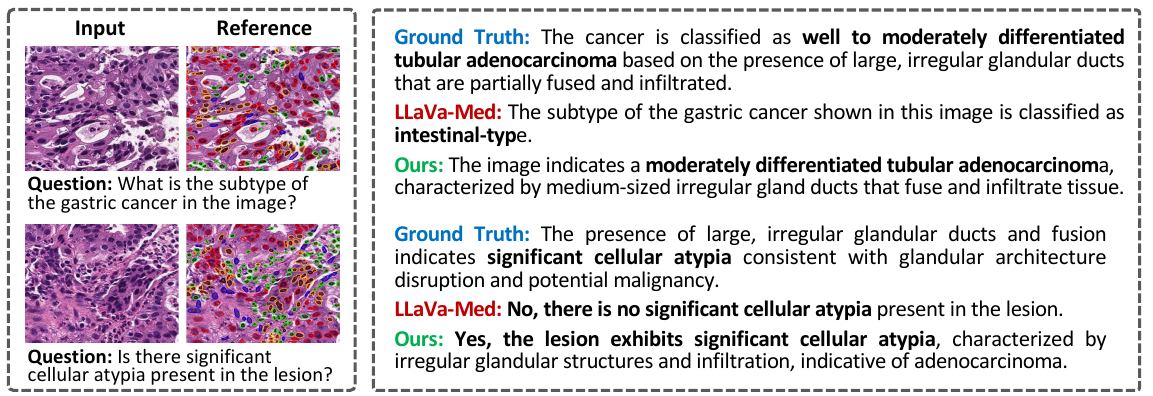}
    \caption{\textbf{Qualitative comparison of conversation task between LLaVA-Med and our proposed PathMR.}  }
    \label{fig6}
\end{figure*}

\begin{table*}[h]
    \centering
    \small
    \renewcommand{\arraystretch}{1.3} 
    \setlength{\tabcolsep}{6pt} 
    \caption{\textbf{Ablation study of module designs on referring segmentation on GADVR dataset.}}
    \begin{tabular}{cc|cc|cc|cc|cc|cc}
        \hline
        \multirow{2}{*}{Cls. Sup.} &  \multirow{2}{*}{Cons.} & \multicolumn{2}{c|}{Neoplastic} & \multicolumn{2}{c|}{Inflammatory} & \multicolumn{2}{c|}{Connective} & \multicolumn{2}{c|}{Epithelial} & \multicolumn{2}{c}{Overall} \\
        \cline{3-12}
        & & gIoU & cIoU & gIoU & cIoU & gIoU & cIoU & gIoU & cIoU & gIoU & cIoU \\
        \hline
        & & 0.655  & 0.736 & 0.474   & 0.503    & 0.474 & 0.498  & 0.240    & 0.300    & 0.527 &  0.634  \\
        \ding{51} &  & 0.659   & 0.740    & 0.476    & 0.500    & 0.474    & 0.502    & 0.253    & 0.310    & 0.531    & 0.644    \\
        \ding{51} & \ding{51} & \textbf{0.672}    & \textbf{0.756}    & \textbf{0.491}    & \textbf{0.513}    & \textbf{0.497}   & \textbf{0.522}    & \textbf{0.264}    & \textbf{0.314}    & \textbf{0.539}    & \textbf{0.657}    \\
        \hline
    \end{tabular}
    \label{tab:sup}
\end{table*}

\subsubsection{Results on Conversation}

To assess the model’s ability in conversation tasks, we evaluate PathMR on both the validation and test sets of the GADVR benchmark. The task involves answering diagnostic questions based on input pathology images, with optional visual references provided for context. We report the quantitative results in Table~\ref{tab:con4}, and provide a qualitative comparison with LLaVA-Med in Fig.~\ref{fig6}.

Our proposed PathMR model achieves the best performance across all metrics, with a BLEU-4 score of 0.301/0.302 and an F1 score of 0.609/0.602 on the validation/test sets, respectively. These results outperform all prior methods, including LLaVA-Med and LISA-13B, both of which rely on large-scale pretraining. 
The qualitative results in Fig.~\ref{fig6} further highlight the superior interpretability and pathology awareness of our model. In both examples, PathMR provides responses that are factually consistent with the visual features in the image, correctly identifying the cancer subtype and the presence of significant cellular atypia. In contrast, LLaVA-Med fails to detect key diagnostic features or gives vague or incorrect answers, such as misclassifying the cancer subtype or overlooking morphological abnormalities.

\subsection{Ablation Studies}
To assess the contribution of key components in our framework, we perform ablation experiments on two modules: class supervision and consistency constraint, with results shown in Fig. \ref{fig8} (reasoning segmentation) and Table~\ref{tab:sup} (referring segmentation). For the reasoning segmentation, incorporating class supervision improves feature discrimination and enhances category-specific predictions, increasing gIoU from 0.602 to 0.609. The consistency constraint alone also provides a notable improvement by stabilizing predictions in spatially fragmented or morphologically ambiguous regions, reaching gIoU 0.605 and cIoU 0.724. When both modules are applied together, the model achieves the best overall performance with gIoU 0.629 and cIoU 0.734, suggesting that both modules work together to enhance the overall performance.

For the referring segmentation task, although the baseline model performs reasonably well, adding class supervision brings clear improvements, particularly in challenging categories such as epithelial cells. Incorporating the consistency constraint further refines the predictions, especially in morphologically diverse regions like inflammatory tissues. When both modules are combined, the method achieves the highest overall performance, with gIoU reaching 0.539 and cIoU reaching 0.657, delivering consistent gains across all tissue types. These results demonstrate that integrating both modules effectively enhances segmentation accuracy and robustness.

\section{Conclusions and Discussion}

In this work, we introduced PathMR, a novel multimodal visual reasoning framework designed to enhance interpretability in computational pathology. By integrating pixel-level segmentation with language-based diagnostic reasoning, PathMR enables transparent and clinically meaningful decision support. Comprehensive experiments on two benchmark datasets demonstrate that PathMR consistently surpasses state-of-the-art visual reasoning models.
Looking ahead, we plan to incorporate tissue-level segmentation labels to improve the spatial localization of nuclei within their anatomical context, enabling more fine-grained and structure-aware diagnostic reasoning. Additionally, we aim to expand PathMR to a broader range of cancer types and tissue morphologies, validating its scalability and robustness across diverse, large-scale datasets. These enhancements will further advance PathMR toward integration into real-world clinical workflows, offering pathologists interpretable, reliable, and evidence-based diagnostic support.

\bibliographystyle{IEEEtran}
\bibliography{tmi}

\end{document}